\pdfoutput=1

\documentclass[11pt]{article}

\usepackage[preprint]{acl}

\usepackage{times}
\usepackage{latexsym}
\usepackage{booktabs}
\usepackage{subcaption}
\usepackage[most]{tcolorbox}
\tcbuselibrary{breakable}

\usepackage[T1]{fontenc}

\usepackage[utf8]{inputenc}

\usepackage{microtype}

\usepackage{inconsolata}

\usepackage{graphicx}
\usepackage{amsmath}
\usepackage{multirow}
\usepackage{xcolor}

\title{Beyond Nash Equilibrium: Bounded Rationality of LLMs and humans in Strategic Decision-making}

\author{
Kehan Zheng\thanks{\ \ Equal contribution.} \quad Jinfeng Zhou\footnotemark[1] \quad Hongning Wang\thanks{\ \ Corresponding author.}\\
The CoAI Group, Tsinghua University \\
    \texttt{\{zhengkh24, zjf23\}@mails.tsinghua.edu.cn} \quad \texttt{hw-ai@tsinghua.edu.cn}
}

\begin{document}
\maketitle
\begin{abstract}

Large language models are increasingly used in strategic decision-making settings, yet evidence shows that, like humans, they often deviate from full rationality. In this study, we compare LLMs and humans using experimental paradigms directly adapted from behavioral game-theory research. We focus on two well-studied strategic games, Rock-Paper-Scissors and the Prisoner's Dilemma, which are well known for revealing systematic departures from rational play in human subjects. By placing LLMs in identical experimental conditions, we evaluate whether their behaviors exhibit the bounded rationality characteristic of humans. Our findings show that LLMs reproduce familiar human heuristics, such as outcome‑based strategy switching and increased cooperation when future interaction is possible, but they apply these rules more rigidly and demonstrate weaker sensitivity to the dynamic changes in the game environment. Model-level analyses reveal distinctive architectural signatures in strategic behavior, and even reasoning models sometimes struggle to find effective strategies in adaptive situations. These results indicate that current LLMs capture only a partial form of human-like bounded rationality and highlight the need for training methods that encourage flexible opponent modeling and stronger context awareness.

\end{abstract}

\section{Introduction}

Strategic decision-making is the process of identifying optimal actions to achieve specific goals. 
This involves calculating optimal solutions based on established rules, predicting the behavior of others from past interactions, and adapting to an evolving environment. \citep{wikipedia_game_theory}. 
As large language models (LLMs) have grown in capability, recent research has begun to investigate their potential as rational agents in strategic settings \citep{Hua2024, Jia2025}. 
However, a consistent finding from empirical studies using game-theoretic and reinforcement-learning frameworks is that LLMs often diverge from theoretically optimal Nash-equilibrium strategies \citep{Akata2025, Hayes2024}. 
This sub-optimality is attributed to several factors, including systematic cognitive biases, misunderstanding of environmental rules, and challenges in updating beliefs or predicting others' actions in evolving environments \citep{Fontana2024, Hayes2024}.

This deviation from rational behavior is not unique to LLMs. 
Human strategic decision-making is also known as "bounded rationality", which describes how cognitive limitations, perceptual filters, and contextual factors cause decisions to diverge from optimal standards \citep{Kahneman1979, Tversky1974}. 
Since LLMs often learn from vast amounts of human-generated data, a critical question emerges: do the patterns of bounded rationality observed in LLMs mirror those of humans? \citep{Guo2024, Cheung2025}. 
The similarity between them reflects LLMs' internalization of strategic reasoning akin to human cognition.
Understanding this connection is vital for guiding the development of LLMs that can adapt more fluidly to real-world strategic interactions and achieve better optimal outcomes \citep{Guo2024, Lyu2025}.
Therefore, a systematic investigation into \textbf{the bounded rationality between LLMs and humans in strategic decision-making} is essential, yet existing research has not adequately addressed this area.

To investigate this question empirically, it is necessary to design a strategic environment that can precisely measure and compare the behaviors of both LLMs and humans, satisfying three core criteria: 
(i) \textbf{a finite action space} to allow for precise quantitative analysis of strategic behaviors; 
(ii) \textbf{a clear, closed-form Nash equilibrium} to serve as a normative benchmark for rational behavior; 
and (iii) \textbf{an extensive behavioral literature documenting human departures from equilibrium play}, which provides a robust empirical baseline for comparison. 
These requirements lead us to two classic strategic games: the zero-sum contest of \textbf{Rock-Paper-Scissors (RPS)} and the non-zero-sum \textbf{Prisoner’s Dilemma (PD)}. 
In RPS, the unique equilibrium is a mixed strategy of selecting each action with equal probability, while in the PD, the Nash equilibrium of mutual defection starkly contrasts with the Pareto-optimal outcome of mutual cooperation. 
Their complementary decision structures provide a robust framework to not only assess whether LLMs approximate optimal strategies but also to examine whether their failures replicate the predictable, bias-driven patterns characteristic of human strategic decision-making \citep{WangXuZhou2014, Hoffman2015, Zhang2021, Ahn2001, Schneider2022}.

We replicate the human-subject experiments of RPS and PD \cite{Zhang2021, DalBo2005}, placing each LLM in the same decision-making environments. 
After logging their action sequences, on the one hand, we compute aggregate performance metrics, such as choice percentages in RPS and cooperation rates in PD, to benchmark against human behavior.
On the other hand, we analyze each LLM individually to understand how factors like model family and reasoning mechanisms influence strategic behavior. 
This two-stage approach provides a comprehensive view of how LLMs align with or diverge from human strategic decision-making.

Our experiments reveal three key findings. First, LLMs adopt human-like decision shortcuts, but in a much more rigid manner. 
Second, LLMs are less adaptive, responding to environmental changes more sluggishly than humans. 
Third, strategic behavior is consistent within model families, creating distinct "strategic signatures". 
Notably, reasoning-enhanced models can solve for the Nash equilibrium in analytically clear games but struggle in scenarios that demand theory-of-mind inference.

Our study introduces a methodology for direct, head-to-head comparisons between human and machine strategic behavior by placing LLMs into human-subject experiments. 
Applying this framework to six models in Rock-Paper-Scissors and the Prisoner’s Dilemma, we evaluate the strategic behavior of LLMs against human players, measuring their bounded rationality and analyzing their decision patterns. 
The findings underscore the need for training methods focused on enhancing opponent modeling and context-aware reasoning to advance LLMs toward human-level strategic adaptability.

\section{Related Work}

\paragraph{Humans' Strategic Behaviors.}

Evidence from behavioral game theory shows that humans consistently diverge from Nash equilibrium. Limited iterative thinking explains non‑equilibrium choices in guessing and matrix games \citep{Nagel1995,STAHL1995218}. Social preferences lead players to sacrifice payoffs for fairness or reciprocity \citep{Rabin1993,FehrSchmidt1999,CharnessRabin2002}. In repeated interactions, costly punishment and intuitive cooperation keep contributions above equilibrium levels \citep{FehrGaechter2000,RandGreeneNowak2012}. Together these results confirm bounded rationality as a regular feature of human play \citep{Camerer1997}.
Classic examples of bounded rationality appear in games like Rock-Paper-Scissors (RPS) and Prisoner's Dilemma (PD). In RPS, human players tend to repeat a move after a victory and switch after a defeat, producing cyclic patterns rather than truly random choices \citep{Dyson2016,WangXuZhou2014}. In iterated PD, cooperation often exceeds the one shot equilibrium prediction: participants use tit for tat, punish defectors and cooperate more when future interaction is likely \citep{DalBo2005,MonteroPorras2022,RandGreeneNowak2012}. These game specific tendencies further underscore bounded rationality in human strategic play.

\paragraph{LLM Performance in Strategic Decision‑making Tasks.}
A rapidly expanding body of work evaluates LLMs in a spectrum of strategic contexts, from simple 2×2 matrix games to rich multi‑agent environments. Existing works have tested models in classic two‑player games, coordination games, public‑goods dilemmas, negotiation and bargaining tasks, large‑scale board games (e.g.\ Diplomacy), and social‑deduction settings (e.g.\ Werewolf) \citep{LoreHeydari2024,Gandhi2023,MetaDiplomacy2022,Hua2024,Ye2025,Mozikov2024,Xia2024,Kwon2024,Deng2024,Bianchi2024,Fish2025}.  This literature shows that while advanced LLMs can plan, cooperate, and even outperform humans under certain conditions, their strategic behavior remains highly sensitive to model family, prompting style, and auxiliary planning modules.
Simple strategic environments with clear Nash equilibria provide a rigorous test of LLM rationality. In the Ultimatum Game, GPT‑4 rejects low but positive offers more often than predicted, accepting only larger splits \citep{Tennant2024,Kwon2024}. In Matching Pennies, models fail to randomize uniformly, showing stable side biases absent explicit randomization prompts \citep{Schneider2023}. Reasoning‑tuned variants such as DeepSeek‑R1 and O1 approach equilibrium play more closely in one‑shot dilemmas, yet still rely on chain‑of‑thought scaffolding to achieve optimal strategies \citep{Brookins2024,Silva2024}, showing that even advanced LLMs exhibit bounded rationality without targeted prompting.

\paragraph{Comparing Human and LLM Behavior in Strategic Decision‑making.}

Recent studies compare language models with humans in three main ways. First, large scale replications give models exactly the same behavioural economics tasks used with laboratory subjects and then match the resulting choice distributions \citep{Mei2024,socialeval}. Second, head‑to‑head designs pair model agents with live human players in repeated games and record relative payoffs and learning curves \citep{Akata2025}. Third, simulation benchmarks check model outputs against documented human heuristics and bias patterns, often varying framing or fairness prompts \citep{Fontana2024,LoreHeydari2024,Hosseini2025,Hagendorff2023,MacmillanScott2024}. Taken together, these work shows that current models often match human averages on cooperation and fairness but still diverge in coordination skill and in how bounded rationality manifests. However, existing studies rarely replicate full sociological experimental protocols, limiting direct comparison under identical conditions. We fill this gap by implementing parallel sociological experiments for both human participants and LLMs, allowing a one to one assessment of their strategic decision making. Moreover, by focusing explicitly on bias and bounded rationality in classic games, we examine whether models and humans exhibit similar systematic deviations from optimal play, thus uncovering the shared and distinct mechanisms underlying their decisions.

\section{Methodology}

To examine whether LLMs exhibit human-like or fully rational patterns of strategic decision-making, we conduct simulation-based experiments modeled after established human-subject protocols. 

\paragraph{Model Selection and Configuration.}

We evaluated six SOTA LLMs from three distinct model families: GPT-4o and O1 \citep{openai2024, openai2024o1}, Claude-3.5 and Claude-3.7 (with extended thinking) \citep{anthropic2024claude35sonnet, anthropic2025claude37sonnet}, and DeepSeek-V3, DeepSeek-R1 \citep{deepseekai2025deepseekv3, deepseekai2025deepseekr1}.
These models are selected to represent a range of reasoning capabilities and alignment strategies, including both general-purpose dialogue agents and models fine-tuned for structured multi-step inference. This diversity allows us to assess how strategic behavior varies across architectures and training regimes.

\paragraph{Experimental Design and Prompting.}

We replicate experimental conditions from the behavioral game theory literature that study human strategic decision-making, adapting these settings for LLMs to enable credible comparisons. When constructing prompts, we closely align the language, incentives, feedback structure, and instructions with those used in human-subject experiments, ensuring consistency in framing and strategic context. 
Each model is queried with a consistent prompt format that includes the current game context, payoff structure, and recent action history. The model is instructed to output both an action (e.g., “Rock” or “Cooperate”) and an explanation of its reasoning.

\paragraph{Comparison and Evaluation.}

Model behavior is evaluated using a set of metrics grounded in game-theoretic principles and established behavioral benchmarks. We compare LLM outputs against theoretical predictions as well as human behavioral patterns, focusing on deviations from optimal strategies and the presence of consistent decision heuristics. To enable meaningful comparison, we apply the same evaluation framework across models and reference human-study datasets. In addition, we conduct model-level analyses to examine behavioral variability across architectures, shedding light on how underlying design influence strategic reasoning.

\section{Experiments}
In this, we first present the experiment setups and results obtained under the two competitive games in Section \ref{sec:RPS} and \ref{sec:PD}, and summarize the findings from the two games in Section \ref{sec:exp-analysis}. 

\subsection{Rock-Paper-Scissors}
\label{sec:RPS}

Rock-Paper-Scissors (RPS) offers a simple yet powerful framework to study competitive decision-making in adversarial settings. In this game, each player simultaneously chooses one of the three actions, rock, paper, or scissors, according to a circular dominance rule: rock beats scissors, scissors beats paper, and paper beats rock. Because no option strictly dominates the others, the game has no pure-strategy equilibrium. Instead, the optimal strategy is to play randomly, choosing each option with an equal probability.

Despite the theoretical prediction, empirical research has shown that human participants struggle to generate truly random sequences. Instead, players often rely on predictable behavioral patterns such as frequency bias or sequential strategies based on past outcomes. One such strategy is Win-Stay/Lose-Change (WSLC), where players repeat a choice after a win and switch after a loss. More complex behaviors have also been documented, including recursive reasoning and one-step-ahead prediction based on an opponent’s historical moves \citep{WangXuZhou2014, Hoffman2015}.

Building on these insights, we investigate whether LLMs exhibit similar behavioral patterns when placed in sequential adversarial settings. In particular, we explore whether LLMs generate action sequences that deviate from randomness, and whether they adopt patterns such as WSLC in response to outcomes. This approach allows us to assess whether LLMs internalize basic principles of adaptive play and whether their decision-making resembles that of human participants.

\paragraph{Setup}

To distinguish between players adopting a purely random strategy and those aligning with a Nash equilibrium strategy, we employ a modified payoff matrix adopted by prior work. In this design, the Nash equilibrium corresponds to a mixed strategy of \( \frac{1}{4} \) for Rock, \( \frac{1}{2} \) for Paper, and \( \frac{1}{4} \) for Scissors, while the uniform random strategy assigns equal probability \( \frac{1}{3} \) to each action.

\begin{table}[t]
\centering
\resizebox{.7\columnwidth}{!}{
\begin{tabular}{lccc}
\toprule
\textbf{} & \multicolumn{3}{c}{\textbf{Player 1}} \\
\textbf{Player 2} & Rock & Paper & Scissor \\
\midrule
Rock     & (2,2) & (1,3) & (4,0) \\
Paper    & (3,1) & (2,2) & (1,3) \\
Scissor  & (0,4) & (3,1) & (2,2) \\
\bottomrule
\end{tabular}}
\caption{The payoff (Player 1, Player 2) matrix table.}
\label{tab:rps-matrix}
\vspace{-3mm}
\end{table}

To define model decision patterns based on the previous round’s choice, we classify each transition into three types: stay (repeat the previous action; e.g., Rock → Rock), upgrade (choose the action that beats the previous one; e.g., Rock → Paper), and downgrade (choose the action that loses to the previous one; e.g., Rock → Scissors).

Firstly, we evaluated each of six LLMs in a fully crossed Rock–Paper–Scissors tournament against all five other models and itself, with each pairing playing 50 rounds.  The primary focus was to identify whether models exhibit frequency bias, sequential dependencies, or simple patterns such as Win‑Stay/Lose‑Change (WSLC).

Secondly, we assessed how LLMs adapt when facing predictable, rule‑based opponents by introducing two bots: one following a Win‑Stay/Lose‑Upgrade (WSLU) algorithm and another following a Win‑Downgrade/Lose‑Stay (WDLS) algorithm.  This setup allows us to probe whether LLMs can detect structured opponent behavior and adjust their responses accordingly.

For full details of match pairings, repetition counts, and transition matrices, see Appendix~\ref{app:experimental_setup}.

\paragraph{Results}

\begin{figure*}[t]
  \centering
  \begin{subfigure}[t]{0.48\textwidth}
    \centering
    \includegraphics[width=\linewidth]{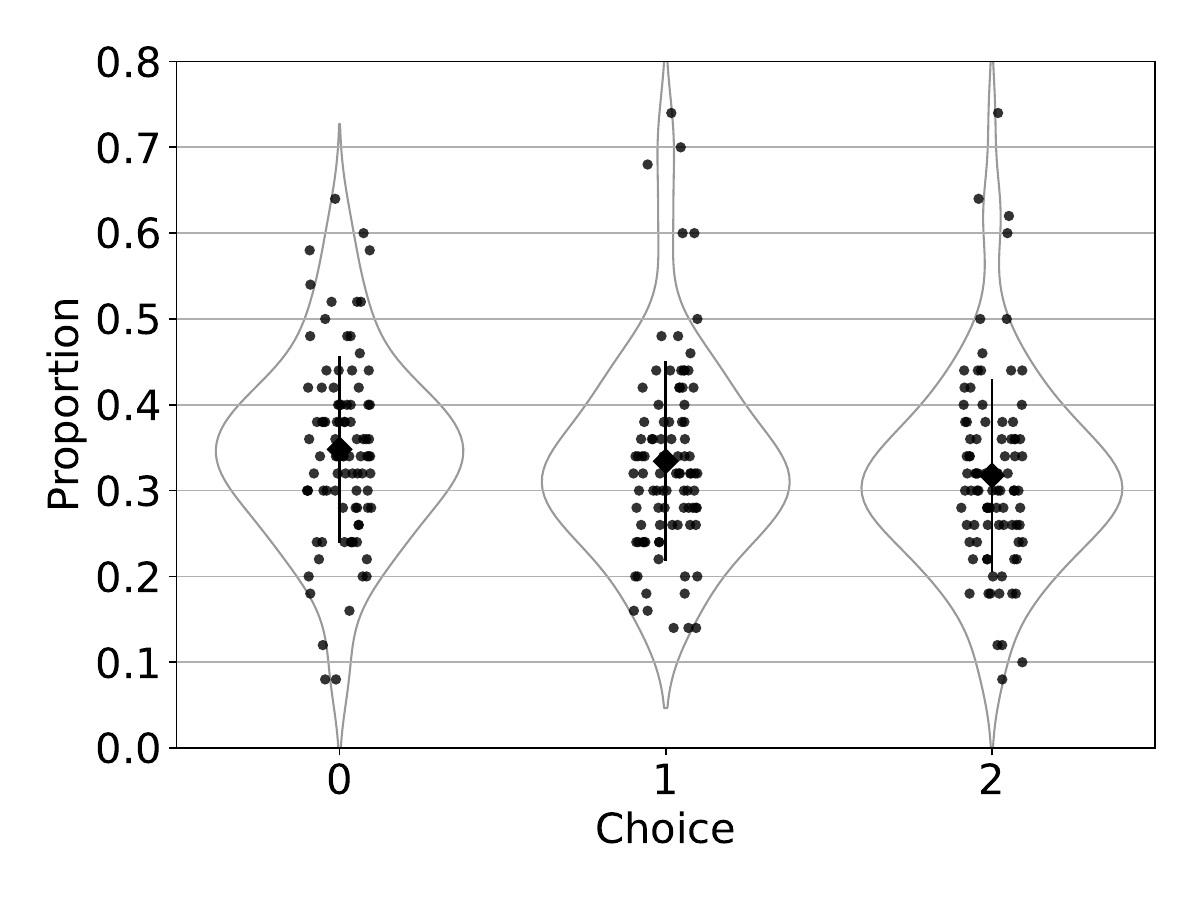}
    \caption{Human choice proportions}
    \label{fig:choice-violin-human}
  \end{subfigure}
  \hfill
  \begin{subfigure}[t]{0.48\textwidth}
    \centering
    \includegraphics[width=\linewidth]{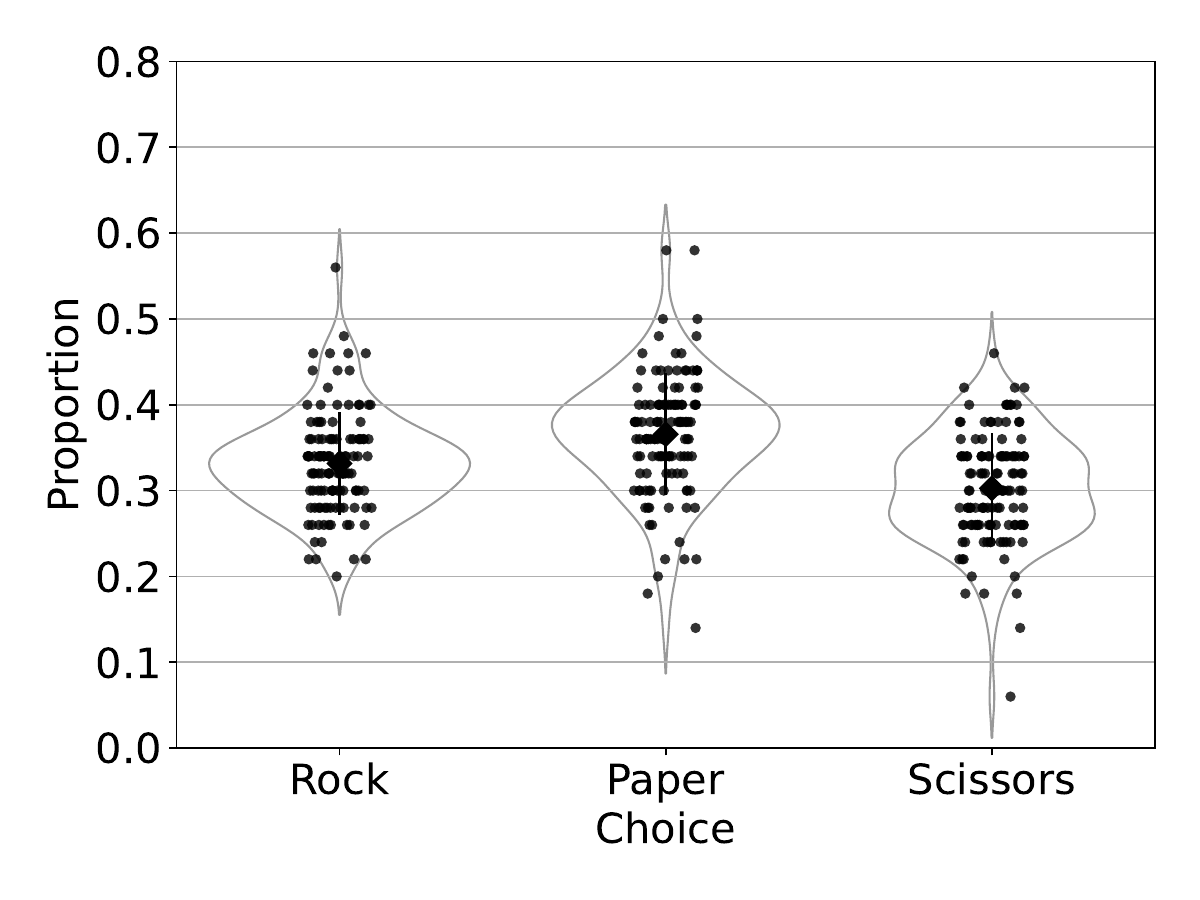}
    \caption{LLM choice proportions}
    \label{fig:choice-violin-llm}
  \end{subfigure}
  \caption{Distribution of Rock–Paper–Scissors choice proportions for (a) human players and (b) LLMs.}
  \label{fig:choice-proportions}
\end{figure*}

For the first stage of the experiment, We examine the proportion of choices made by each agent to assess whether their behavior aligns with the theoretical Nash Equilibrium or follows a purely random strategy. The choice distribution plots are shown as Figure~\ref{fig:choice-proportions}.

\begin{figure*}[t]
  \centering
  \begin{subfigure}[t]{.8\textwidth}
    \centering
    \includegraphics[width=\linewidth]{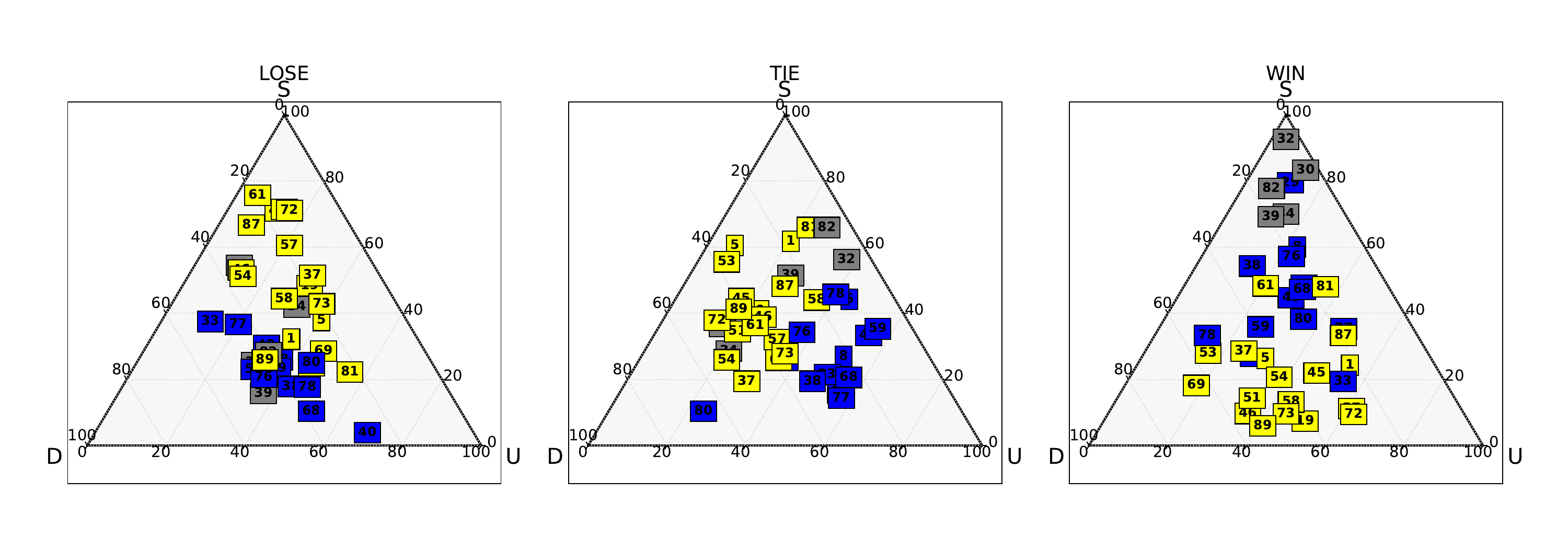}
    \caption{Human, outcome‐based}
    \label{fig:ternary-human-cond}
  \end{subfigure}
  
  \begin{subfigure}[t]{.8\textwidth}
    \centering
    \includegraphics[width=\linewidth]{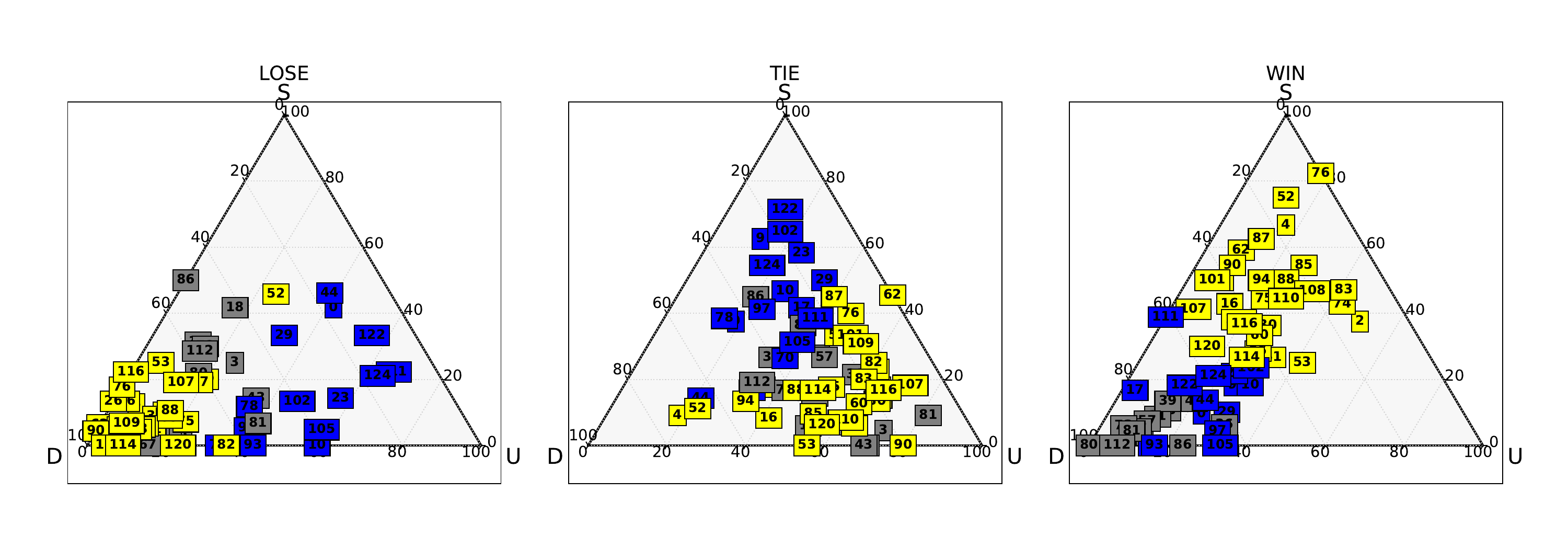}
    \caption{LLM, outcome‐based}
    \label{fig:ternary-llm-cond}
  \end{subfigure}
  \caption{Ternary plots of strategy proportions following different outcomes for outcome‐based agents.}
  \label{fig:ternary-outcome}
\end{figure*}

To assess whether players’ next choices depend on the outcome of the previous round (win, loss, or tie), we conduct chi-square tests of independence. Among human participants, 36 in 90 show significant outcome‑action dependence; among model runs, 63 in 126 do. To further clarify individual decision patterns, we cluster the outcome-based participants into three representative strategy categories. Figure \ref{fig:ternary-outcome} reports the ternary distributions of strategies (stay, upgrade, downgrade) by different outcomes for outcome-based humans and LLMs. 

\begin{figure*}[t]
  \centering
  \begin{subfigure}[t]{0.49\textwidth}
    \centering
    \includegraphics[width=\textwidth]{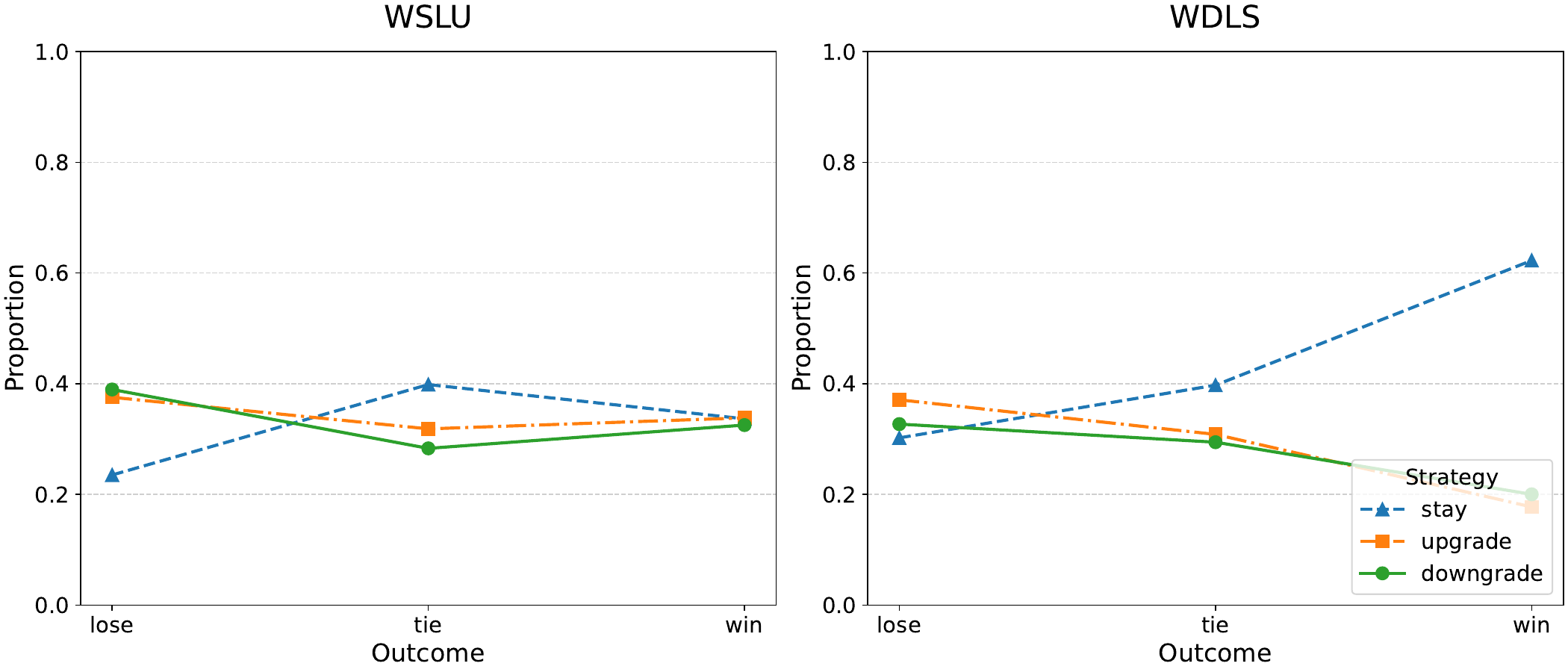}
    \caption{Humans under WSLU vs.\ WDLS}
    \label{fig:strategy-distribution-human}
  \end{subfigure}
  \hfill 
  \begin{subfigure}[t]{0.49\textwidth} 
    \centering 
    \includegraphics[width=\textwidth]{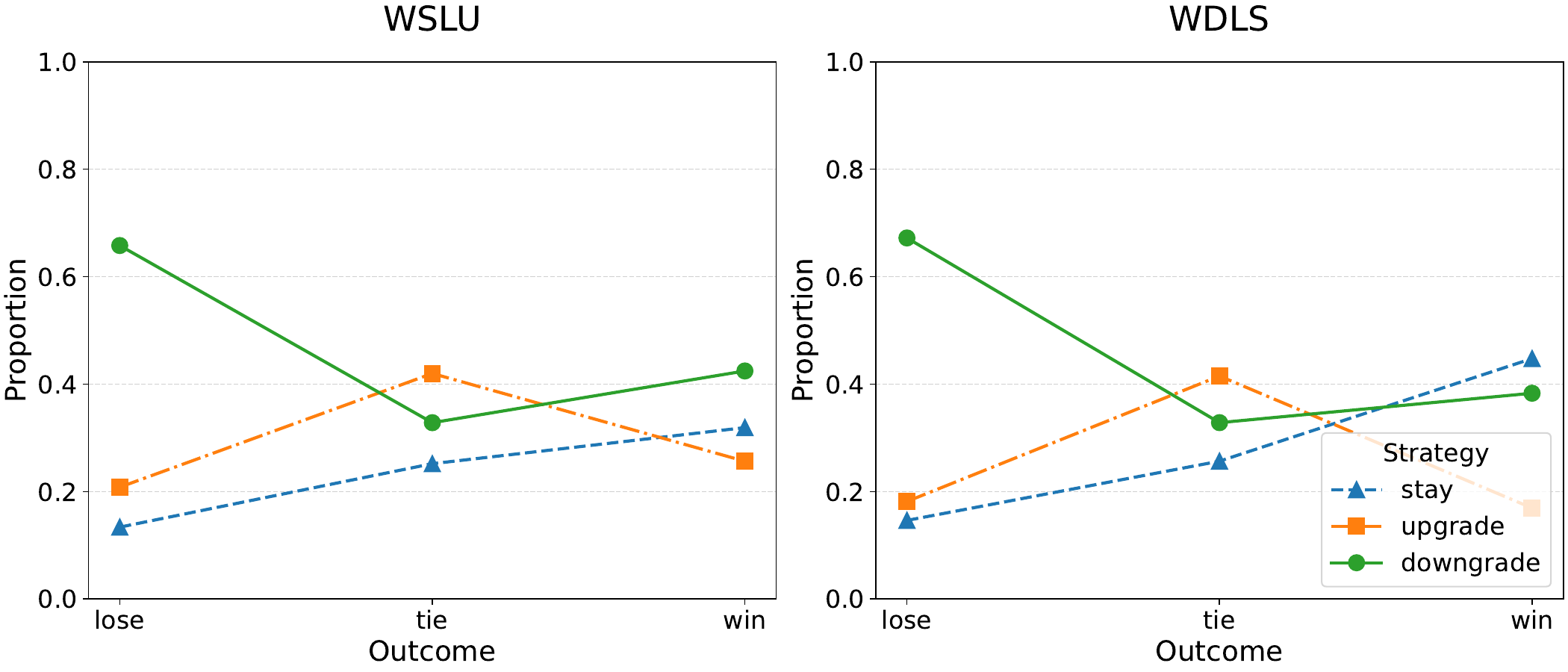}
    \caption{LLMs under WSLU vs.\ WDLS}
    \label{fig:strategy-distribution-llm}
  \end{subfigure}
  \caption{Comparison of outcome‐conditioned strategy distributions under WSLU vs.\ WDLS policies. (a) Human players. (b) LLM players.}
  \label{fig:strategy-distribution-all}
\vspace{-3mm}
\end{figure*}

In the second stage of our experiment that pairs agents with WSLU and WDLS bot, we summarize their win counts and cumulative payoffs in Table~\ref{tab:exp2-performance}, and visualize their strategy distribution under each condition in Figure~\ref{fig:strategy-distribution-all}.

\begin{table}[t]
\centering
\resizebox{\columnwidth}{!}{
\begin{tabular}{llcc}
\toprule
\textbf{Opponent Strategy} & \textbf{Agent Type} & \textbf{Win Differential} & \textbf{Payoff Differential} \\
\midrule
\multirow{2}{*}{WSLU} 
 & Human & +2.01 & +6.30 \\
 & LLM   & +4.50 & +14.56 \\
\midrule
\multirow{2}{*}{WDLS} 
 & Human & +6.42 & +24.72 \\
 & LLM   & $-$6.44 & $-$15.56 \\
\bottomrule
\end{tabular}}
\caption{Average win and payoff differentials (Agent $-$ Bot) by opponent strategy and agent type. Positive values indicate that the agent outperformed the bot.}
\label{tab:exp2-performance}
\end{table}

\begin{figure*}[t]
  \centering
  \begin{subfigure}[t]{0.32\textwidth}
    \centering
    \includegraphics[width=\linewidth]{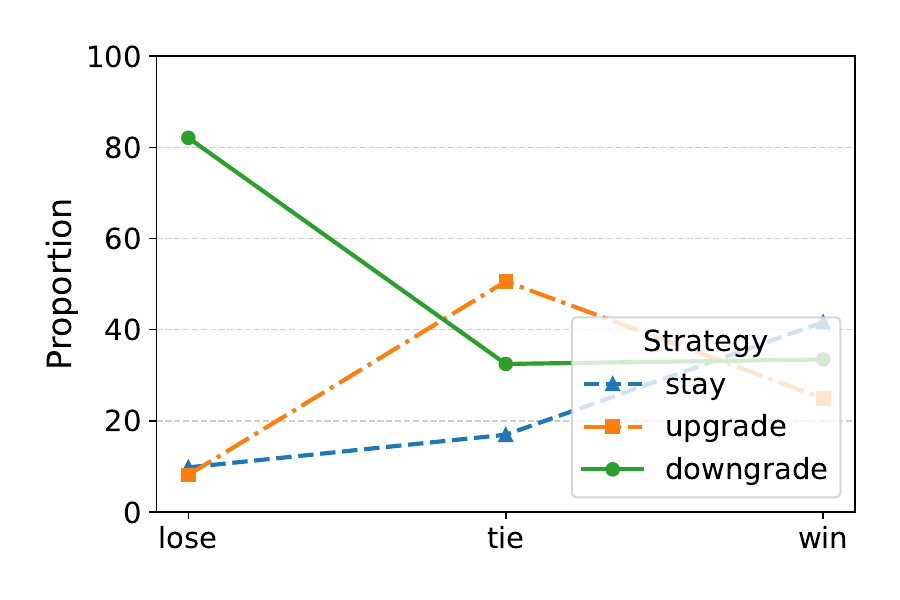}
    \caption{GPT‑4o}
    \label{fig:strategy-gpt4o}
  \end{subfigure}\hfill
  \begin{subfigure}[t]{0.32\textwidth}
    \centering
    \includegraphics[width=\linewidth]{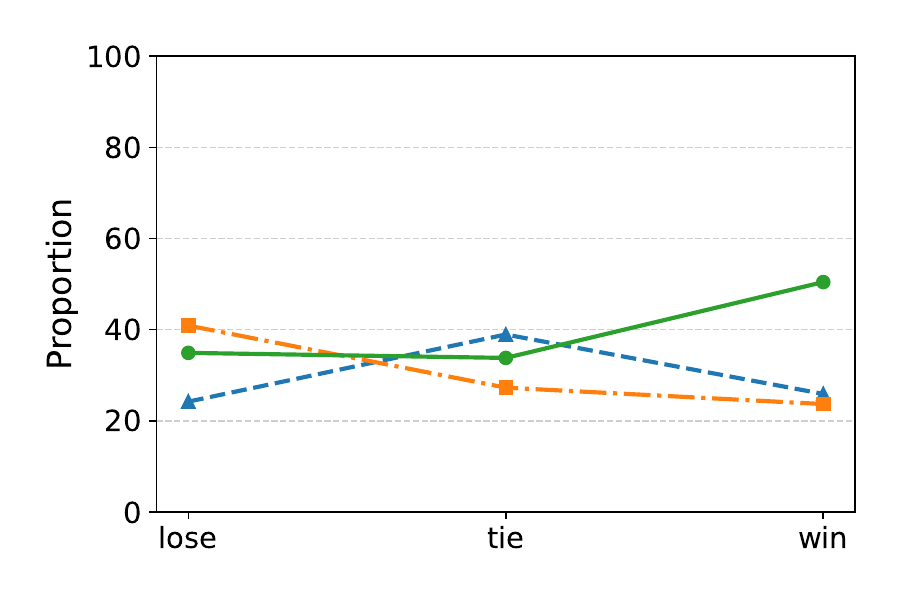}
    \caption{DeepSeek-V3}
    \label{fig:strategy-ds-chat}
  \end{subfigure}\hfill
  \begin{subfigure}[t]{0.32\textwidth}
    \centering
    \includegraphics[width=\linewidth]{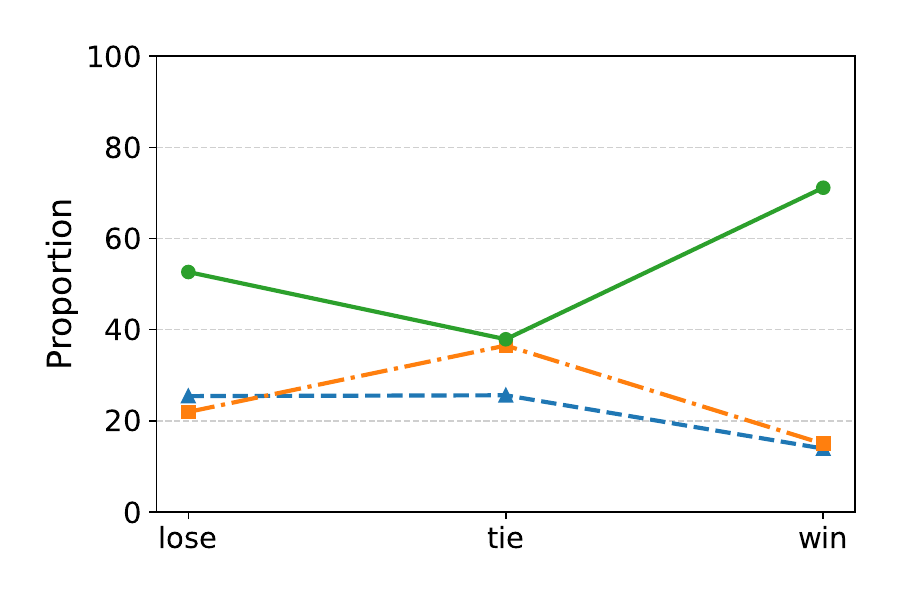}
    \caption{Claude 3.5}
    \label{fig:strategy-claude35}
  \end{subfigure}

  \vspace{1em} 

  \begin{subfigure}[t]{0.32\textwidth}
    \centering
    \includegraphics[width=\linewidth]{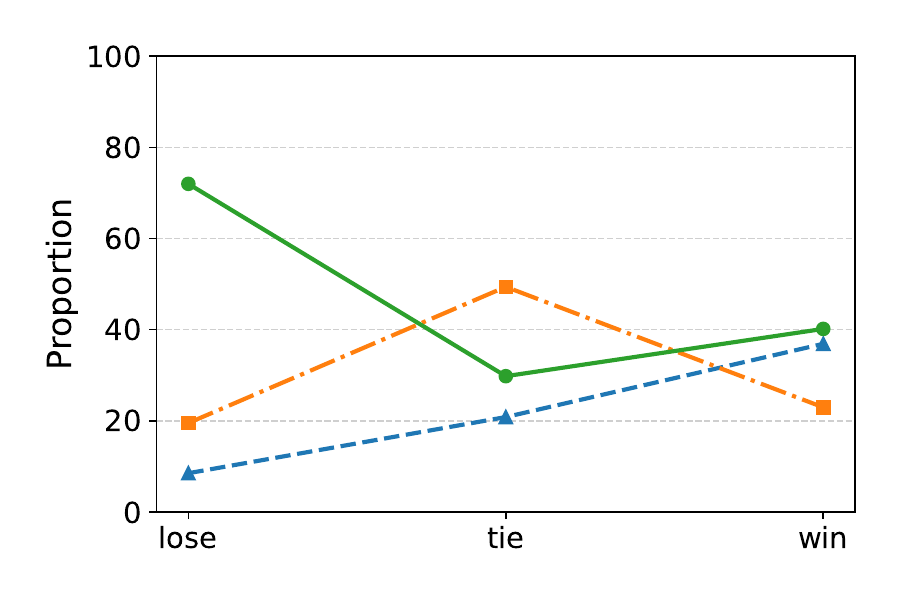}
    \caption{O1}
    \label{fig:strategy-o1}
  \end{subfigure}\hfill
  \begin{subfigure}[t]{0.32\textwidth}
    \centering
    \includegraphics[width=\linewidth]{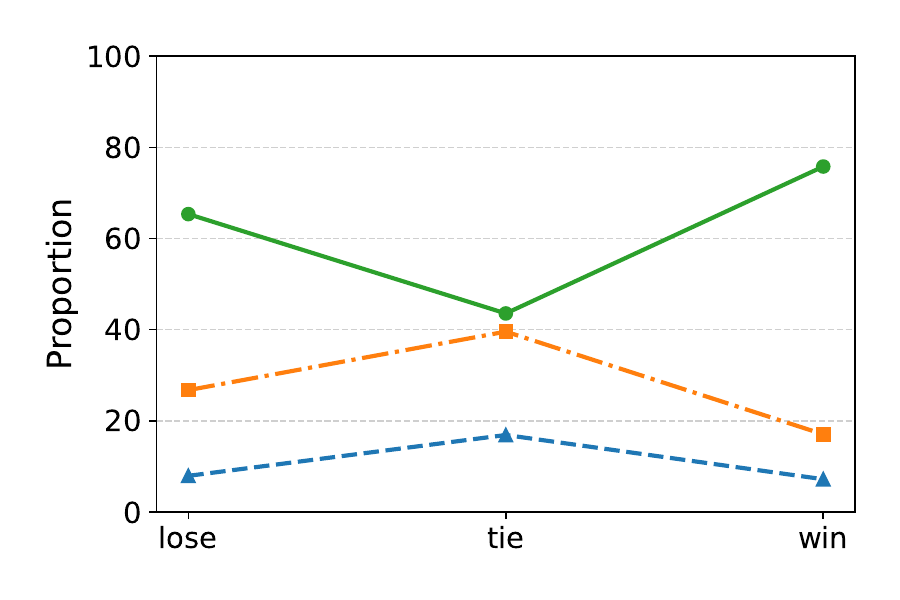}
    \caption{DeepSeek-R1}
    \label{fig:strategy-ds-reasoner}
  \end{subfigure}\hfill
  \begin{subfigure}[t]{0.32\textwidth}
    \centering
    \includegraphics[width=\linewidth]{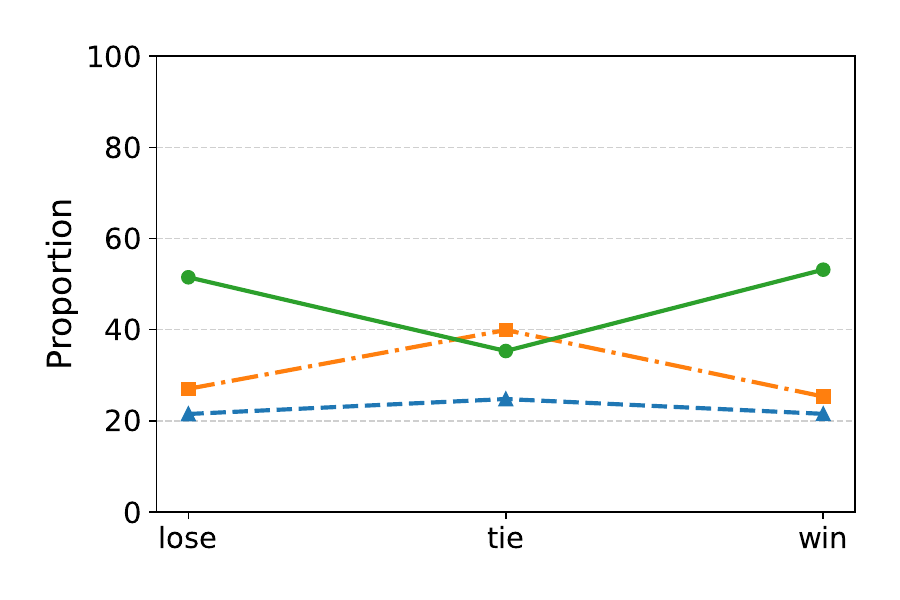}
    \caption{Claude 3.7}
    \label{fig:strategy-claude37}
  \end{subfigure}

  \caption{Outcome‐conditioned strategy distributions, grouped by model class.
    (a) GPT‑4o, (b) DeepSeek-V3, (c) Claude-3.5, (d) O1, (e) DeepSeek-R1, (f) Claude-3.7.}
  \label{fig:strategy-per-model-class}
\vspace{-3mm}
\end{figure*}

\begin{table}[t]
\centering
\resizebox{\columnwidth}{!}{
\begin{tabular}{lrrrr}
\toprule
\textbf{Model} & \multicolumn{2}{c}{\textbf{WSLU}} & \multicolumn{2}{c}{\textbf{WDLS}} \\
\cmidrule(lr){2-3} \cmidrule(lr){4-5}
 & Win Diff Avg & Payoff Diff Avg & Win Diff Avg & Payoff Diff Avg \\
\midrule
GPT‑4o       & $\,12.33$ & $\,36.00$  & $-6.33$  & $-23.33$  \\
DeepSeek‑V3  & $ -3.67$  & $ -8.67$   & $-5.67$  & $-15.33$  \\
Claude‑3.5   & $\,0.00$  & $\,2.67$   & $-1.00$  & $ -7.33$  \\
O1           & $\,5.33$  & $\,24.00$  & $-6.00$  & $ -8.00$  \\
DeepSeek‑R1  & $\,7.00$  & $\,15.33$  & $-20.33$ & $-52.67$  \\
Claude‑3.7   & $\,6.00$  & $\,18.00$  & $\,0.67$ & $\,13.33$ \\
\bottomrule
\end{tabular}}
\caption{Average win and payoff differentials by model under WSLU and WDLS conditions.}
\label{tab:exp2-differentials-combined}
\vspace{-3mm}
\end{table}

To investigate decision patterns at the model level, we build on the previous experiments in two settings. In the LLM vs LLM matches, we compute the outcome‑conditioned transition frequencies for each model. In the LLM vs bot trials, we record each model’s win count difference and cumulative payoff difference against the WSLU and WDLS bots. The results are respectively presented in Figure~\ref{fig:strategy-per-model-class} and Table~\ref{tab:exp2-differentials-combined}.
\subsection{Prisoner's Dilemma}
\label{sec:PD}

The Prisoner’s Dilemma is a classic example of a social dilemma in which two players must simultaneously choose between two options: to cooperate or to defect. In theory, the optimal strategy under finite rounds is to defect, as defection strictly dominates cooperation in each individual round. However, empirical studies involving human participants have consistently shown that cooperation is not only possible but also fairly common, even in one-shot or finite-round versions of the game. \citep{Ahn2001, Schneider2022}

One key factor that may help explain the gap between theoretical predictions and observed human behavior is the possibility of future interaction. Game theorists have long acknowledged that repeated play and the expectation of continued engagement can profoundly influence strategic choices. When individuals anticipate future encounters, they often develop implicit systems of punishment and reward, which help deter opportunistic behavior and foster cooperation \citep{DalBo2005}.

Building on this foundation, it is worthwhile to explore how LLMs behave under similar conditions in the Prisoner’s Dilemma. By simulating strategic environments modeled after human studies \citep{DalBo2005}, we aim to examine whether and how LLMs adjust their behavior in response to the potential for future interaction. This comparison offers an opportunity to investigate the extent to which model behavior aligns with or diverges from human decision-making in social dilemmas.

\begin{table}[t]
\centering
\resizebox{.5\columnwidth}{!}{
\begin{tabular}{lll}
\hline
 & \textbf{L} & \textbf{R} \\
\hline
\textbf{U} & 65 , 65 & 10 , 100 \\
\textbf{D} & 100 , 10 & 35 , 35 \\
\hline
\end{tabular}}
\caption{\label{tab:pd-matrix}
Payoff matrix for Prisoner's Dilemma. 
}
\vspace{-3mm}
\end{table}

\paragraph{Setup}

The design of the LLM experiment follows the logic of the human study, with adjustments to accommodate the nature of AI agents.
We employ two session structures in this experiment:  
\emph{Dice} sessions inform the model that after each round the game continues with probability $\delta$, while  
\emph{Finite} sessions specify a fixed horizon $H$ in the prompt. 
Each PD session runs all three treatments concurrently: Dice with continuation probabilities $\delta\in\{0,0.5,0.75\}$ and Finite with fixed horizons $H\in\{1,2,4\}$, where $H=1/(1-\delta)$.  By matching the expected number of rounds across Dice and Finite treatments, we ensure a fair comparison of LLM decision patterns under equivalent temporal incentives.
Dice sessions embed an explicit “shadow of the future,” with $\delta$ controlling the strength of expected continuation, while Finite sessions present a known, fixed horizon without uncertainty.  This contrast isolates the impact of future expectations on model cooperation and defection behavior.  
For full details of continuation schedules, pairing algorithms, and prompt templates, see Appendix~\ref{app:experimental_setup}.

\paragraph{Results}

The primary outcome measure in our study is the \textit{cooperation rate} per round, defined as the proportion of cooperative choices made by agents in each round of the repeated Prisoner's Dilemma game. Tables~\ref{tab:overall-coop} and~\ref{tab:overall-coop-llm} summarize the average cooperation rates observed in the human and LLM experiments. 
We compute, for each LLM its overall Prisoner’s Dilemma cooperation rate in every experimental condition as shown in Table~\ref{tab:llm-coop-modelwise}.

\begin{table}[t]
\centering
\resizebox{\columnwidth}{!}{
\begin{tabular}{lllccl}
\hline
\multicolumn{2}{c}{\textbf{Dice}} & & \multicolumn{2}{c}{\textbf{Finite}} & \\
\cline{1-2} \cline{4-5}
\textbf{Treatment} & \textbf{Cooperation (\%)} & & \textbf{Treatment} & \textbf{Cooperation (\%)} & \\
\hline
$\delta = 0$   & 9.17  & & $H = 1$ & 10.34 & \\
$\delta = 0.5$ & 27.41 & & $H = 2$ & 10.11 & \\
$\delta = 0.75$& 37.64 & & $H = 4$ & 21.43 & \\
\hline
\end{tabular}}
\caption{\label{tab:overall-coop}
Percentage of cooperation by treatment in human study.
}
\vspace{-4mm}
\end{table}
\subsection{Analysis}
\label{sec:exp-analysis}

After collecting the behavioural data from the RPS and PD experiments, building on the metrics in human study, we first contrast model outputs with human benchmarks to probe bounded rationality. We then drill down to model–family differences to uncover how architecture and alignment shape strategic tendencies. The analysis is organised around two research questions.

\begin{table}[t]
\centering
\resizebox{\columnwidth}{!}{
\begin{tabular}{lllccl}
\hline
\multicolumn{2}{c}{\textbf{Dice}} & & \multicolumn{2}{c}{\textbf{Finite}} & \\
\cline{1-2} \cline{4-5}
\textbf{Treatment} & \textbf{Cooperation (\%)} & & \textbf{Treatment} & \textbf{Cooperation (\%)} & \\
\hline
$\delta = 0$   & 33.33 & & $H = 1$ & 16.15 & \\
$\delta = 0.5$ & 38.37 & & $H = 2$ & 21.35 & \\
$\delta = 0.75$& 37.83 & & $H = 4$ & 23.96 & \\
\hline
\end{tabular}}
\caption{\label{tab:overall-coop-llm}
Percentage of cooperation by treatment in LLM experiments.
}
\vspace{-3mm}
\end{table}

\begin{table}[t]
\centering
\resizebox{\columnwidth}{!}{
\begin{tabular}{llcccccc}
\hline
\textbf{Type} & \textbf{Treatment} & GPT‑4o & DeepSeek‑V3 & Claude‑3.5 & o1 & DeepSeek‑R1 & Claude‑3.7 \\
\hline
\multirow{3}{*}{Finite}
 & $H=1$     & 21.9 & 12.5 & 37.5  & 0.0 & 0.0 & 25.0 \\
 & $H=2$     & 37.5 & 25.0 & 25.0  & 3.1 & 0.0 & 37.5 \\
 & $H=4$     & 46.9 & 21.1 & 30.5  & 3.1 & 1.6 & 40.6 \\
\hline
\multirow{3}{*}{Dice}
 & $\delta=0$   & 31.2 &  3.1 & 100.0 & 3.1 & 0.0 & 62.5 \\
 & $\delta=0.5$ & 48.4 & 22.7 & 100.0 & 9.0 & 1.6 & 65.2 \\
 & $\delta=0.75$& 36.6 & 13.5 &  98.7 & 4.3 & 0.8 & 65.9 \\
\hline
\end{tabular}}
\caption{\label{tab:llm-coop-modelwise}
Percentage of cooperation by LLM model, experiment type, and treatment.}
\vspace{-3mm}
\end{table}

\subsubsection*{RQ1: Do LLMs replicate human patterns of bounded rationality?}

\paragraph{LLMs mirror but amplify human heuristics.}  
Across both games, models show the same qualitative departures from Nash equilibrium that typify human play, yet the magnitude of those deviations is larger. In RPS, both humans and LLMs select Rock, Paper, and Scissors with near-equal frequencies instead of Nash Equilibrium (Figure~\ref{fig:choice-proportions}). However, LLMs show more concentrated choice distributions, while human participants more often show skewed preferences. Outcome-based analysis reinforces the trend: a certain percentage of humans and LLMs exhibit statistically outcome dependence in their choices. However, models show more stereotyped response patterns, particularly a strong preference for lose–downgrade transitions, which dominate post-loss moves in many models, often exceeding 60\%. By contrast, humans exhibit diverse outcome-based patterns, including both win–stay/lose–change and win–change/lose–stay strategies, reflecting multiple coexisting heuristics rather than a single dominant bias (Figure~\ref{fig:ternary-outcome}).
In PD, both humans and models increase cooperation under infinite-horizon treatments, consistent with the “shadow of the future” effect (Tables~\ref{tab:overall-coop} and~\ref{tab:overall-coop-llm}). Yet, LLMs display an exaggerated sensitivity: even when \(\delta=0\), a condition theoretically equivalent to the one-shot game (\(H=1\)), models show notably higher cooperation than in the finite counterpart. It shows LLMs may be overly influenced by the presence of future-oriented phrasing in prompts, generalizing cooperative heuristics beyond the structural incentives observed in humans.

\paragraph{LLMs show weaker environmental sensitivity.}  
Humans flexibly adapt heuristics to opponent structure, boosting win–stay by over \(20\) percentage points when facing the downgrade‑oriented WDLS bot (Figure~\ref{fig:strategy-distribution-human}).  LLMs show only minimal adjustment with their dominant lose‑downgrade behavior remaining unchanged (Figure~\ref{fig:strategy-distribution-llm}). This corresponds to humans outperforming both WSLU and WDLS bots, whereas LLMs beat WSLU by +4.50 wins but lose to WDLS by –6.44 wins (Table~\ref{tab:exp2-performance}). A parallel pattern appears in PD: humans raise cooperation from \(\delta=0.5\) (27.4\%) to \(\delta=0.75\) (37.6\%), whereas LLMs actually dip slightly from 38.4\% to 37.8\% (Tables~\ref{tab:overall-coop} and~\ref{tab:overall-coop-llm}), showing that models respond to the presence of a future but not to increases in continuation probability.

\paragraph{Implications.}  
The findings show that while LLMs adopt core bounded‑rational heuristics found in humans, such as outcome‑driven adjustments and future-driven cooperation, they do so in a more rigid, exaggerated form. Models exhibit fixed, stereotyped biases (e.g. lose–downgrade and baseline cooperation) instead of the heterogeneous mix of strategies seen in human players, and they are markedly less responsive to environmental cues like opponent rules or varying continuation probabilities. 

\subsubsection*{RQ2: How does model architecture influence strategic behaviour?}

\paragraph{Model families exhibit distinct strategic signatures.}  
Outcome‑conditioned transitions in LLM-LLM RPS games (Figure~\ref{fig:strategy-per-model-class}) reveal distinct architectural signatures: GPT‑4o and O1 downgrade after a loss in more than 70\% of cases, while Claude‑3.5 and Claude‑3.7 display a win‑downgrade loop, preferring to downgrade even after victories. In PD (Table~\ref{tab:llm-coop-modelwise}), only the Claude family sustains consistently high cooperation, whereas all other models remain below the 50\% mark.

\paragraph{Reasoning models excel only under explicit equilibrium conditions.}  
Reasoning variants excel when a unique optimum is analytically clear. In the one‑shot PD, DeepSeek‑R1 and O1 defect in almost all rounds, matching the Nash prediction, while GPT-4o and DeepSeek-V3 still cooperate for a few percentage (Table~\ref{tab:llm-coop-modelwise}). The picture changes when adaptation to opponent structure is required. Against the WSLU and WDLS bots in RPS, O1 and DeepSeek‑R1 secure only moderate gains under WSLU (+24.00 and +15.33) but suffer substantial losses under WDLS (-8.00 and -52.67) (Table~\ref{tab:exp2-differentials-combined}), offering no clear advantage over general models. Thus, reasoning models approach optimality in games with explicit solutions but lag when success hinges on theory‑of‑mind and flexible opponent modeling.

\paragraph{Implications.}  
LLMs' strategic behavior is tightly bound to architectural lineage. Family‑level biases indicate that pre‑training and alignment choices hard‑wire distinct decision styles. Reasoning models achieve textbook equilibrium when the solution is analytically explicit but fail to leverage theory‑of‑mind in rule-based adversarial settings. This gap highlights the importance of enhancing reasoning pipelines with richer opponent modeling and explicit theory‑of‑mind scaffolds so that strategic adaptation becomes as flexible as human play, rather than limited to problems with closed‑form solutions.

\section{Discussion}

We adapt human studies of Rock–Paper–Scissors and Prisoner’s Dilemma into text‑based trials for six leading LLMs, using identical payoffs, instructions and evaluation metrics. By replaying full sociological protocols rather than isolated prompts, we directly compare model behavior against large human datasets.

According to our analysis, while LLMs share key bounded-rational heuristics with humans, rigidity and reduced sensitivity to context in their application suggest that LLMs exhibit only a partial, amplified form of human‑like bounded rationality rather than a fully analogous strategic reasoning profile.  

Moreover, we conduct a model‑level analysis that reveal two insights: (1) Each model family carries a distinct strategic “fingerprint”, with pre‑training and alignment choices imprinting stable decision styles across both games. (2) Reasoning models outperform general models when the optimal move is analytically obvious, but struggle to infer and adapt to opponents’ strategies, highlighting a persistent gap in theory‑of‑mind–driven reasoning.

Closing that gap will require more than scale or chain‑of‑thought prompts. We recommend integrating opponent‑aware fine‑tuning objectives, reinforcement learning on diverse human gameplay traces and multi‑stage prompting schemes that explicitly invoke theory‑of‑mind. Such hybrid strategies hold promise for endowing LLMs with the flexible, socially aware reasoning that characterizes human decision makers.

Our work adapts established human experimental protocols for RPS and PD into a unified framework for both LLMs and human participants, runs extensive trials under identical conditions, and systematically evaluates strategic choices using the same metrics.  We further dissect model‑family and reasoning‑tuned behaviors to map out enduring decision patterns and highlight areas where adaptive strategic decision-making remains limited.

\section{Limitations}

Our study has several limitations. First, LLM behavior is highly sensitive to prompt design, and minor changes in wording may influence decisions in unpredictable ways. Second, while we replicate human experimental settings, the text-based interaction mode of LLMs lacks the social and contextual richness of human play. Third, we focus on two specific games (RPS and PD), which may limit the generalizability of our findings to broader strategic environments. Fourth, the number of LLMs evaluated is limited, and future work should incorporate a broader range of models to strengthen the generality of observed patterns.

\appendix

\section{Experimental details}
\label{app:experimental_setup}

\subsection{LLM Data Collection}
We evaluate six state‑of‑the‑art LLMs via their official APIs using identical prompts. The models and their names specified in APIs are as follows: GPT‑4o (GPT-4o-2024-11-20), O1 (o1-2024-12-17), DeepSeek‑V3 (deepseek-chat), DeepSeek‑R1 (deepseek-reasoner), Claude‑3.5‑Sonnet (claude-3-5-sonnet-20241022), and Claude‑3.7‑Sonnet (claude-3-7-sonnet-20250219, with extended thinking).

To preserve behavioral variability and allow for probabilistic action distributions, we set the decoding temperature to 1.0 for all models. This high-temperature setting ensures that the models do not collapse to deterministic outputs and can express stochastic strategies when appropriate.

\subsection{Rock-Paper-Scissors}
\paragraph{Setup.} 
In the first part, we examine how LLMs behave in an environment with minimal structure and no explicit opponent modeling. Each LLM plays repeated Rock-Paper-Scissors games for 50 rounds against six types of opponents: itself and each of the other five models. Every pairing is repeated three times to ensure robustness. This setup results in \( \binom{6}{2} \times 3 = 45 \) distinct matches, forming a fully crossed interaction network. The payoff matrix is shown in Table~\ref{tab:rps-matrix}.

The second part builds on prior human-subject studies that investigate how players adapt when facing structured, rule-based opponents. We use two bot opponents: one following a \textit{Win-Stay/Lose-Upgrade} (WSLU) strategy, and the other following a \textit{Win-Downgrade/Lose-Stay} (WDLS) strategy. Both bots operate probabilistically, with an 80\% probability for their primary behavior after each outcome and 10\% for each alternative action. In tie situations, actions are selected uniformly at random among the three options: stay, upgrade, and downgrade. The full transition probabilities for each strategy are presented in Table~\ref{tab:bot-strategy-matrix}.

\begin{table}[t]
\centering
\resizebox{\columnwidth}{!}{
\begin{tabular}{llccc}
\toprule
\textbf{Algorithm} & \textbf{Outcome} & \textbf{Stay} & \textbf{Upgrade} & \textbf{Downgrade} \\
\midrule
\multirow{3}{*}{WSLU} 
 & Win  & 0.8  & 0.1  & 0.1 \\
 & Tie  & 0.33 & 0.33 & 0.33 \\
 & Lose & 0.1  & 0.8  & 0.1 \\
\midrule
\multirow{3}{*}{WDLS} 
 & Win  & 0.1  & 0.1  & 0.8 \\
 & Tie  & 0.33 & 0.33 & 0.33 \\
 & Lose & 0.8  & 0.1  & 0.1 \\
\bottomrule
\end{tabular}}
\caption{Probabilistic strategies used by the two bot opponents in the second experiment. Values indicate the probability of selecting Stay, Upgrade, or Downgrade after each type of outcome.}
\label{tab:bot-strategy-matrix}
\end{table}

Each LLM is paired with both WSLU and WDLS bots for 50 rounds, repeated three times.

\paragraph{Prompts.} All prompts (exemplified) employed in this experiment are shown as below.

\begin{tcolorbox}[
    colback=white,
    colframe=black,
    width=0.48\textwidth,    
    breakable,
    left=2mm,                
    right=2mm,               
    top=1mm,                 
    bottom=1mm               
]
[\textbf{System Message}] \\
Rock-Paper-Scissors \\
You have been randomly paired with a computer algorithm (i.e., opponent) to play the Rock-Paper-Scissors game. You and your opponent will make decisions at the same time across multiple trials. In each trial, each of you will have to simultaneously select one of three options: Rock, Paper, or Scissors. The outcome of a trial (Lose, Win, or Tie) will depend on the decisions that both you and your opponent make according to these basic rules of the game: \\
-- Rock beats Scissors \\
-- Scissors beats Paper \\
-- Paper beats Rock \\
Your payoff in a given trial will depend on the decisions that both you and your opponent make. **Beating the opponent brings you more points than tying (choosing the same option as the opponent), which brings you more points than if you get beaten by your opponent**. The exact payoffs for all possible outcomes are fixed throughout the
game. After each trial, information will be provided to both you and your opponent about what both of you did and what were the corresponding payoffs in the previous trial. \\
Your payment will be given according to **the sum of points that you accumulate across all trials, as advertised, plus a base payment for successful completion of the study.** \\

[\textbf{Decision Message}] \\
Trial 6 \\
Please make a choice: Rock, Paper, or Scissors. Think before you make your choice. \\
Output with the following format: \\
Reason: [Your reason for the choice] \\
Choice: [Rock/Paper/Scissors] \\

[\textbf{Feedback Message}] \\
Feedback in the previous trial: \\
You lost! \\
You chose Paper, your opponent chose Scissors. \\
Your payoff was 1, your opponent's payoff was 3. \\
Your total payoff so far is 42. Your opponent's total payoff so far is 42.
\end{tcolorbox}

\subsection{Prisoner's Dilemma}
\paragraph{Setup.}
We consider two types of session structures as below.

Dice: The model is informed that after each round, the game may continue with a fixed probability $\delta \in \{0, 0.5, 0.75\}$. The game terminates according to a simulated draw from a Bernoulli process. The continuation probability is made explicit to the model in its prompt.

Finite: The total number of rounds is known in advance and specified in the model’s prompt. We use $H \in \{1, 2, 4\}$, matching the expected lengths of the infinite sessions. This enables a controlled comparison of decision patterns between sessions with and without a 'shadow of the future'.


To control for learning effects from one treatment to another, two sessions are run for each kind of continuation rule (Dice and Finite). For example, for Dice we run one session with the order \((\delta = 0,\, \delta = 0.5,\, \delta = 0.75)\) and another with the inverse order \((\delta = 0.75,\, \delta = 0.5,\, \delta = 0)\).  We call the first kind of session “Normal” and the last kind “USD” (up-side-down).

A rotation matching scheme is used to avoid potential interaction and contagion effects between different repeated games. In each session, subjects are divided into two groups: \textit{Red} and \textit{Blue}. In each match, every Red subject is paired with a distinct Blue subject.

Given that each subject is matched only once with each subject from the opposite group, the total number of matches in a session is \(\frac{N}{2}\), where \(N\) is the number of subjects in a session. Since there are three treatments per session, each treatment contains \(\frac{N}{6}\) matches.

For experiments conducted on LLMs, we set \(N = 24\), with each LLM type participating as 4 distinct agents in a game. This configuration allows for up to 4 matches per treatment per subject, ensuring sufficient coverage and interaction diversity across treatments.

\paragraph{Prompts.} All prompts (exemplified) employed in this experiment are shown as below.

\begin{tcolorbox}[
    colback=white,
    colframe=black,
    width=0.48\textwidth,    
    breakable,
    left=2mm,                
    right=2mm,               
    top=1mm,                 
    bottom=1mm               
]
[\textbf{System Message}] \\
You are an UCLA undergraduate recruited for a research project. \\
The session you are participating in is broken down into 3 separate parts. At the end of  the last part, you will be paid the total amount you have accumulated during the course of the 3 parts in addition to the show-up fee. \\
Your dollar earnings at the end of the session are determined by the points/\$ exchange rate posted on the board in the  front and back of the room. This exchange rate is equal to 200 points/\$. Therefore, 200 points are equivalent to \$1. \\
The participants are divided in two groups: Red and Blue.  \\
Red and Blue participants will be matched together to interact in the following way. The Red participant can choose between U or D and the Blue participant can choose between L and R.  \\
If the Red participant chooses U and the Blue participant chooses L, both earn 65 points.\\
If the Red participant chooses U and the Blue participant chooses R, the Red participant earns 10 and the Blue participant earns 100 points. \\ 
If the Red participant chooses D and the Blue participant chooses L, the Red participant earns 100 and the Blue participant earns 10 points.  \\
If the Red participant chooses D and the Blue participant chooses R, both earn 35 points. \\
The points of the Red participants are indicated on the screen in red, and the Blue participant points are indicated in blue.  \\
We will show the result of previous rounds of the current match.\\
Remember that you are a Red participant. \\

[\textbf{Dice-mode introduction}] \\
We will begin the first part now. This first part will consist of 4 matches. In each match every Red participant is paired with a Blue participant. You will not be paired twice with the same participant during the session or with a participant that was paired with someone that was paired with you or with someone that was paired with someone that was paired with someone that was paired with you, and so on. Thus, the pairing is done in such a way that the decisions you make in one match cannot affect the decisions of the participants you will be paired with in later matches or later parts of the session. \\
In this part, after each round we will roll a four sided dice. If the numbers 1, 2 or 3 appear, the participants will interact again without changing pairs. If a 4 appears, the match ends and participants are re-matched to interact with other participants. Therefore, in this part, each pair will interact until a 4 appears. After that, a new match will start with different pairs. Therefore you will interact until a 4 appears, with 4 different participants.  \\
You will now participate in 4 matches, each match paired with a different participant. In each match you will interact with the same person until a 4 appears. Remember: your decisions in one match cannot affect the decisions of the people you will interact with in future matches. This is not a practice; you will be paid! \\
If you are a Red participant you can choose the actions in  red, U or D, and if you are a Blue participant you can press the actions in Blue, L or R. \\

[\textbf{Choice Message (New match)}] \\
You are now matched with a new participant. You will interact with this participant until a 4 appears. Make your choices now. \\
Think before you make your choice. \\
Output with the following format: \\
Reason: Your reason for the choice \\
Choice: U/D \\

[\textbf{Feedback Message}] \\
Feedback in the previous rounds: \\
Your choices: U  \\
Opponent choices: L \\
Your total payoff: 65 \\
Opponent total payoff: 65 \\

[\textbf{Dice Message (Continue)}]
A 2 appeared therefore this match continues. Now you are in Round 2 of the same match. You are still interacting with the same participant. You can see in the history the result of the previous rounds. Make your choices now.
Think before you make your choice. \\
Output with the following format: \\
Reason: Your reason for the choice \\
Choice: U/D \\

[\textbf{Dice Message (End)}]
A 4 appeared therefore this match ended. You have earned 65 points. Now you will be matched with the next participant.

\end{tcolorbox}

\end{document}